# Evaluating Complex Task through Crowdsourcing: Multiple Views Approach

Lingyu Lyu
Department of CECS
University of Louisville
Louisville, Kentucky, USA
l0lv0002@louisville.edu

Mehmed Kantardzic
Department of CECS
University of Louisville
Louisville, Kentucky, USA
mehmed.kantardzic@louisville.edu

*Abstract*—With the popularity of massive open online courses (MOOCs), grading through crowdsourcing has become a prevalent approach towards large scale classes. However, for getting grades for complex tasks, which require specific skills and efforts for grading, crowdsourcing encounters a restriction of insufficient knowledge of the workers from the crowd. Due to knowledge limitation of the crowd graders, grading based on partial perspectives becomes a big challenge for evaluating complex tasks through crowdsourcing. Especially for those tasks which not only need specific knowledge for grading, but also should be graded as a whole instead of being decomposed into smaller and simpler sub-tasks. We propose a framework for grading complex tasks via multiple views, which are different grading perspectives defined by experts for the task, to provide uniformity. Aggregation algorithm based on graders' variances are used to combine the grades for each view. We also detect bias patterns of the graders, and de-bias them regarding each view of the task. Bias pattern determines how the behavior is biased among graders, which is detected by a statistical technique. The proposed approach is analyzed on a synthetic data set. We show that our model gives more accurate results compared to the grading approaches without different views and de-biasing algorithm.

## I. INTRODUCTION

Grades and comments from instructors are very important to evaluate the level of the students' understanding and provide guidance for future studying process. Traditional ways of grading by instructors or teaching assistants are becoming a big challenge for large classes such as massive open online classes (MOOCs), which are distributed on platforms such as Udacity, Coursera and EdX [1], and the grading processes are really time consuming and tedious. As an example, there were around 100,000 students signed up for the online machine learning course offered on the Coursera platform. To evaluate students' work, a scalable way for grading is required for such large scale classes. Due to the restriction of un-scalability of the traditional way, grading through crowdsourcing has been developed as one of the prevalent ways to solve the problem.

Since Amazon launched Mechanical Turk in 2005, crowdsourcing has become a powerful mechanism for completing large scale tasks which involve human intelligence computing. The tasks in MTurk range from labeling images with keywords to writing product descriptions. However, the types of tasks accomplished through MTurk have typically been limited to those that are low in complexity, independent, and require little time and cognitive effort to complete [15]. In contrast to the typical tasks posted on MTurk, there are many tasks that are complex and need specialized skills to deal with in the real world. Students' essays or research papers are examples of complex tasks which are different from traditional MTurk's tasks.

To explore the use of crowdsourcing in complex task grading which is not decomposable into sub-tasks, consider for example the usual experience of giving course evaluations. The evaluations are given by students to reflect the success of the teachers' teaching skills for the whole semester. It actually is also a task of grading. Usually, every student in one class is asked to give course evaluation at the end of the semester. It includes questions such as: did the instructor explain concepts correctly; did the instructor provide clear constructive feedback; was the course effectively organized, etc. After each question, the evaluation scale is provided for the student to pick from. This evaluation process of instructors' success in teaching the course is a grading task through crowdsourcing, where students are used as a crowd. In order to give the evaluation for the course, it is necessary to provide grades considering the full semester, instead of decomposing the task into weekly or monthly based evaluation. Another issue arises during the evaluating process is that different students would grade based on their own perspectives and experiences without any guidance information. To ensure perspective uniformity, those instructive questions are provided, which are also called views, for the students grade. After getting grades for each view, an overall grade could be obtained by aggregating the view grades. Derived from this idea, we propose the approach of crowdsourcing grading process through different views for complex tasks such as evaluating programming projects or research papers.

The idea makes use of different views, which are diverse grading perspectives worked as instructive information to provide uniformity among graders. These different views are specified by experts before distributing the tasks to the graders for evaluation. We describe the graders used in the complex task evaluation as experienced domain grader, which means they have some knowledge about the domain of the assignment, but at the same time are different from experts in the field. Insufficient knowledge of the graders may lead to the existence of bias in the grading process regarding each view. By detecting the bias behavior of the graders, it is possible to de-bias them statistically, and significantly improve the quality of the grading process.

There are several concepts used in this work, and they are defined as follows:

*Experts* – Those who have enough knowledge and ability to have best understanding of the complex task, e.g. instructors, who specified and assigned the complex assignment for students, are considered as experts.

*Experienced domain grader* – Those who have some knowledge about the domain of the assignment, but at the same time are different from experts that they are not as knowledgeable and experienced in the field. For example, the students, who have taken and successfully finished the course, could be counted as an experienced domain grader for corresponding course assignment.

*Complex task* – The task which cannot be simply decomposed into smaller sub-tasks, but instead is able to define different views for it. As an example, grading essays or computer science projects can be viewed as a complex task.

*Observed view grade* – Observed view grade is the grade given by for every view by each grader without any processing.

*Consensus view grade* – The grade we get for each view after applying aggregating algorithms or some other processing algorithms, which used to combine observed view grades from different graders for the same submission.

*Overall grade* – The grade we get after combing all the view grades according to the instructions of the expert. It is the grade for every submission instead of each view.

*True grade* – The grade which could reflects the true quality of the submission of the complex assignment. True grades are the grades given by experts.

We present in this paper the framework for crowdsourcing a grading process for complex tasks. Two main contributions of our approach are: extending the Vancouver algorithm [2], proposed by L. de Alfaro, and are applied for different views of complex tasks defined by an expert. Applying de-biasing process to discover biased graders, and the biases are removed from the final grades.

The paper is organized as follows: Section II gives the literature review of current methods used for grading by crowdsourcing. Section III describes the proposed Vancouver-based method with de-biasing. Section IV shows our experiment results. Section V presents comparison of the out-come of this work with other methodologies.

## II. PREVIOUS WORK

### 2.1 Literature Review

To relieve the grading burden of the experts in the traditional way [20-22], especially with the popularity of open online classes such as MOOCs, peer reviewing has been proposed as a new way to address the grading tasks [1-2][23-25]. As stated in [24], "students are trained to be competent reviewers and are then given the responsibility of providing their classmates with personalized feedback on expository writing assignments". Most of the peer evaluation systems proposed nowadays are similar to this concept. In [1], three statistical models are developed for estimating true grades in peer grading. The models put prior distributions over latent variables such as true scores, grader's reliability and bias. Gibbs sampling [13] and expectation maximization (EM) [6-12] is then used as approximate inference approach to estimate the true score. A system called CrowdGrader is developed in [2] to explore the use of peer feedback in grading assignments. It introduced a Vancouver algorithm (derived from [14]) which relies on a reputation system to aggregate the peer reviewing grades. [23] applied Bayesian data analysis to model a computer supported peer review process in a legal class. In [24], a web-based peer review program called Calibrated Peer Review (CPR) system is introduced to improve student learning. Peer reviewing experiments have shown promise, but it may also cause problems. As what Richard Smith proposed in his paper, "the practice of peer review is based on faith in its effects, rather on facts" [3]. Suppose all the students in one class try to give good grades to each other, then with the absence of the true grades provided by the instructor or teaching assistant used for comparison, the consensus grades would be the submissions' grades. However, these grades actually cannot give correct estimation of the true grades.

Many complex tasks, including complex grading tasks, are difficult to be done by crowdsourcing through MTurk due to its typical limitations. In [15] a framework was proposed to solve the problem by breaking down the complex task into a sequence of subtasks. Then subtasks are done through MTurk. This idea inspired by MapReduce [16] systems consists of three steps: partition, map and reduce. [17-19] presented the similar approaches to address the problem encountered while crowdsourcing the solutions to complex tasks. In [17], crowdsourcing is used as a novel approach to grade the math through Internet, by splitting the expert task to non-expert. Divide-and-conquer algorithm, which depicts the 'decompose, solve and recompose' structure, is proposed in [18] to solve general problem via crowdsourcing. In [19], a system called PlateMate is introduced to crowdsource nutritional analysis from photographs via MTurk. Foods in each image are identified and measured separately. Although crowdsourcing a complex task by decomposing it into subtasks offers tradeoff between cost and performance, the problem arises when the complex task is un-decomposable. The framework is no longer available when the complex task should be solved as a whole.

Mostly, bias is unavoidable when collecting the data from crowdsourcing. Bias may be caused by personal preference, systematic misleading, and lack of interest [26]. Many researches have included biases analysis in crowdsourcing models to improve the approximation accuracy [26-28]. In [26], a Bayesian model, named as Bayesian Bias Mitigation for Crowdsourcing (BBMC), is proposed to capture the sources of bias. Authors in [27] introduced and evaluated probabilistic models that can detect and correct task-dependent biases in crowdsourcing automatically. In [1], statistical models have been presented to infer the graders' biases in peer reviewing process.

### 2.2 Vancouver Algorithm

The Vancouver algorithm [2] measures each student's grading accuracy, by comparing the grades assigned by the student with the grades given to the same submission by other students in the crowd. It gives more weight to the input of students with higher measured accuracy.

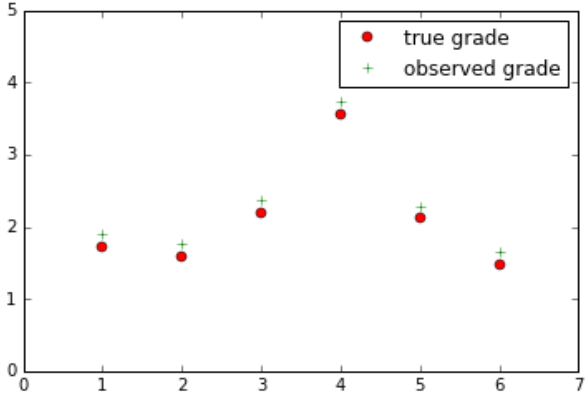 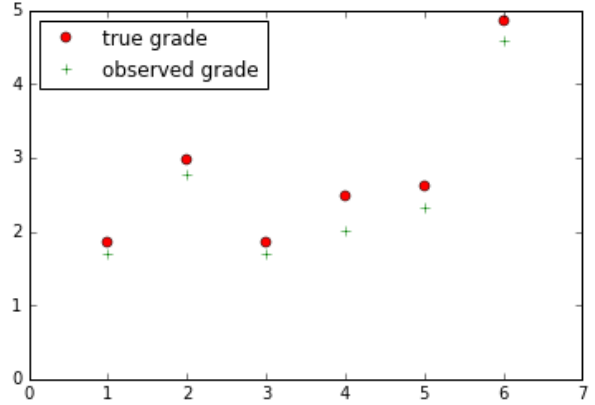

(a) Positive Bias Pattern  (b) Negative Bias Pattern

Fig. 1 Bias Patterns

The Vancouver algorithm is based on the fact of variance minimization principle. This principle said that we could weigh the input provided by student i in proportion to 1/$v_i$, where $v_i$ is grader i's variance. The algorithm proceeds in iterative fashion, using consensus grades to estimate the grading variance of each user, and using the information on user variance to compute more precise consensus grades. The approach may be formalized as: We denote by U the set of students, and by S the set of items to be graded (the submissions). We let G = (T, E) be the graph encoding the review relation, where T = S ∪ U and S ∩ U = ∅, and where (i, j) ∈ E iff j reviewed i; for (i, j) ∈ E, we let $g_{ij}$ be the grade assigned by j to i. We denote by ∂t the 1-neighborhood of a node t ∈ T.

The algorithm proceeds by updating estimates $v_j$ of the variance of user j ∈ U, and estimates $c_i$ of the consensus grade of item i ∈ S, and estimates $v_i$ of the variance with which ci is known. To produce these estimates, the algorithm relies on messages m = (l, x, v) consisting of a source l ∈ S ∪ U, of a value x, and of a variance v. We denote by $M_i$, $M_j$ the lists of messages associated with item i ∈ T or user j ∈ U. Given a set M of messages, we indicate by:

$$E(M) = \frac{\sum_{(l,x,v)\in M} x/v}{\sum_{(l,x,v)\in M} 1/v} \quad (1)$$

$$var(M) = \left(\sum_{(l,x,v)\in M} \frac{1}{v}\right)^{-1} \quad (2)$$

The best estimator $E$(M) we can obtain from M, and its variance $var$(M).

## III. CROWDSOURCING THE GRADING PROCESS OF COMPLEX TASKS THROUGH DIFFERENT VIEWS

The Vancouver algorithm is extended in order to aggregate the observed view grades into consensus view grades. Section 3.1 describes the modified approach. Bias is analyzed for graders regarding each view, for the complex task here. Section 3.2 depicts how the bias is evaluated through bias patterns, and the way to de-bias the graders. Section 3.3 gives the details of the crowdsourcing process.

*3.1 Modified Vancouver Algorithm*

After defining multiple views for the grading tasks, each view will be considered as one separate aggregation task, and the modified Vancouver algorithm is applied to iteratively estimate the consensus grades for views. The extension of Vancouver algorithm is still based on the same principle of Vancouver algorithm, the main difference is that multiple views are taken into account.

The details of the modified Vancouver approach can be seen in Algorithm 1. S is the set of submissions need to be graded, and U is the set of graders. $M_i, M_j$ are the lists of messages associated with submission $i$ and grader $j$. The set of views and observed view grades are used as the input for the algorithm, as well as the review graph. Instead of propagating only one variance as in basic Vancouver algorithm, different variances are considered for multiple views of each grader in the extended approach. After getting consensus view grades, it is required to combine the view grades into overall grade for each submission. The combining methods depends on the instructions of the experts: One way is just simply add the view grades up to get the overall consensus grade for the submission. Another way could be given the weights of each view of the assignment from the expert, aggregating the view grades in proportion to their weights. The extended Vancouver algorithm incorporates the consensus view grades combining steps as in lines 40-45 in Algorithm 1. Here, we assume each view is given a weight associated to it, thus the consensus grade for each submission can be obtained by adding up view grades times their weight.

*3.2 De-biasing*

Bias patterns may be identified by analyzing the graders' fluctuation around true grade, for each view of the complex task. Bias is then reduced from each view grade for the biased graders with detected patterns.

The bias pattern is defined as: for all the submissions graded by the grader g, if most of his/her grades for view v is consistently higher (or lower) than the true grades of the view, then we say the g has a bias pattern on v for the assignment. There are two different types of bias pattern: a) most of the grader's scores of the view are higher than the true score, we call this pattern as "positive bias pattern", and b) graders' scores are lower than the true score, which we call "negative bias pattern". Figure 1 gives an example for each of these two bias patterns. X-axis represents different submissions graded by the grade (for one view) and y-axis is the score scale. For the specific grader,

```
_________________________________________________
             Algorithm 1 Modified Vancouver Algorithm
_________________________________________________
Input: A review graph G = ((S ∪ U), E) such that |∂t| > 1 for all t ∈ S ∪ U, along
with {g_ij[view]}(i,j)∈E, and number of iterations K > 0, views set V and $view \in V$.
Output: Estimates $\hat{q}_i[v]$ for i ∈ S.
1: {Initialization}
2: for all i ∈ S do
3:     for all view ∈ V do
4:         M_i[view] := {(j, g_ij[view] , 1) | (i, j) ∈ E}.
5:     end for
6: end for
7: for iteration k = 1, 2, . . . , K do
8:     {Propagation from submissions}
9:     for all j ∈ U do
10:        for all $view \in V$ do
11:            M_j[view] := ∅
12:        end for
13:    end for
14:    for all i ∈ S do
15:        for all j ∈ ∂i do
16:            for all $view \in V$ do
17:                Let M_¬j[view] = {(j·, x, v) ∈ M_i[view] | j· ≠ j} in
M_j[view] := M_j[view] ∪ (i, E(M_¬j[view] ), var(M_¬j[view] ))
18:            end for
19:        end for
20:    {Propagation from graders}
21:    for all i ∈ S do
22:
22:        for all $view \in V$ do
23:            M_i[view] := ∅
24:        end for
25:    end for
26:    for all j ∈ U do
27:        for all i ∈ ∂j do
28:            for all $view \in V$ do
29:                Let M_¬i[view] = {i· , (x − g_i·j)² , v) | (i· , x, v) ∈ M_j, i·≠ i }
in M_i := M_i ∪ (j, g_ij, E(M_¬j ))
30:            end for
31:        end for
32:    end for
33: end for
34: {Final Aggregation}
35: for all i ∈ S do
36:    for all $view \in V$ do
37:        $\hat{q}_i[view]$ := E(M_i)
38:    end for
39: end for
40: for all $i \in S$ do
41:    $\hat{q}_i := 0$
42:    for all $view \in V$ do
43:        $\hat{q}_i := \hat{q}_i + \hat{q}_i[view] \times weight[view]$
44:    end for
45: end for
_________________________________________________
```

his/her grades given by '+' sign are consistently higher (or lower) than true grades represented by small red circles.

We propose the approach to detect and measure bias pattern as the percentage of values that lie within a width of two standard deviations around the mean of the differences of the true grades and observed grades. Below we present the detailed way how we recognize the bias pattern.

Suppose there are n different submissions graded by grader g, for each of the views of the assignment v. True grades are given as T, where $t_i \in T$ is the true grade of submission i for this view. O is the set of observed grades from g, where $o_i \in O$ is the grade given by g to *i*th submission for v. Denote diff($t_i$, $o_i$) as the difference for each pair of $t_i$ and $o_i$. Calculate:

$$\mu = \frac{1}{n}\sum_{i=1}^n diff(t_i, o_i) \qquad (3)$$

$$\sigma^2 = E((O - \mu)^2) \qquad (4)$$

$$(\mu - 2\sigma, \mu + 2\sigma) \qquad (5)$$

Interval (5) represents the band around the mean of the differences of the true grades and observed grades with a width of two standard deviations. In statistics, we could make sure approximately 95% difference values would fall into the range as stated in (5). That means, if $\mu - 2\sigma$ as calculated from equation (3) and (4) is greater than 0, then we could say 95% of the difference values ($diff(t_i, o_i)$) are positive. In other words, 95% of the grader's grades for view v are higher than the true grades. We state that this grader has a positive bias pattern with 95% confidence for view v. Similarly, if the result of $\mu + 2\sigma$ is negative, we have 95% confidence that the grader has a negative bias pattern for view v. Grades from the graders with negative bias pattern or positive bias pattern are then de-biased to give a better performance before applying aggregating algorithm.

The way to de-bias the observed view grades provided by the biased graders varies on different occasions. The approaches for de-biasing might be subtracting: a) min($diff(t_i, o_i)$); b) max($diff(t_i, o_i)$); c) median of ($diff(t_i, o_i)$) or d) average of ($diff(t_i, o_i)$) from the observed view grades. The best method would be determined based on experimental heuristics.

*3.3 Grading Complex Task through Crowdsourcing*

This research develops a crowdsourcing methodology for grading complex tasks which will result in consensus grades as close as possible to expert grades. By defining different views for the complex task by expert, the approach provides uniformity among the graders crowd. The framework of the approach is presented in Fig. 2 and explained as follows:

1) Define different views for the complex task: Different views are first defined for the complex task by experts. As an example, while grading a research paper, the instructor may define views for the assignment such as a) comprehensiveness; b) enough quotes; c) examples, and inferences; d) technical strength; e) data sets selection and f) presentation, etc. It is also required that an expert gives the scale for each view of the assignment such as 0 to 5 for background information, 0 to 10 to technical strength, and 0 to 3 to data sets selection, etc.

2) Distribute the submissions to graders: each submission should be reviewed and graded by several graders, and each grader is able to review and grade several submissions. The experienced domain graders are assumed in the proposed approach. In addition, we assume that each grader must review a minimum number of submissions to ensure that each submission would have enough reviews. The problem may occur in real world applications: if there are graders who do not do their reviews, which will lead to insufficient number of reviews for some submissions. An online algorithm is proposed in [2] to solve the defect. The algorithm dynamically estimates the probability that each review task will be completed, on the basis of previous history, and assigns review tasks to achieve

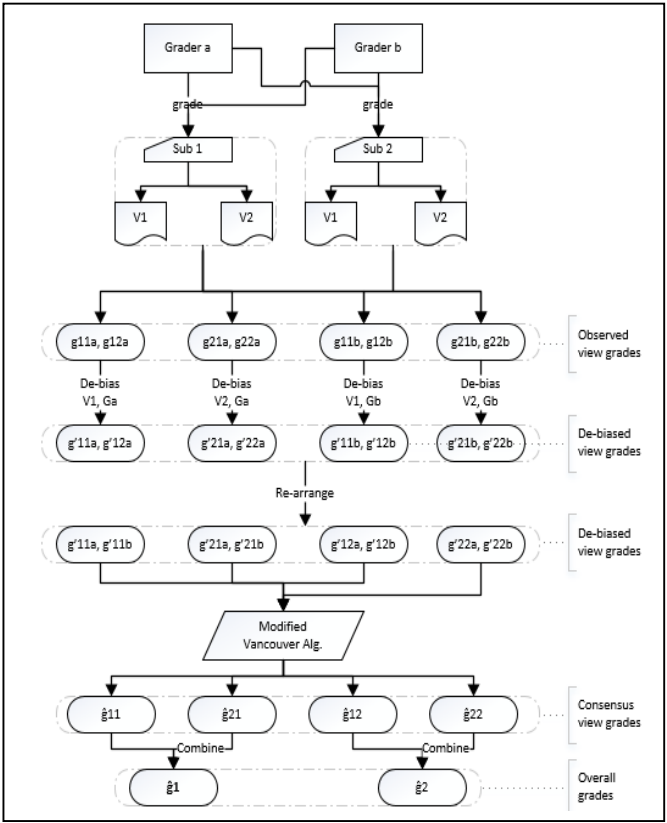

Fig. 2 Framework of Grading Complex Task through Crowdsourcing (Assuming two graders are employed, and two views are defined for the task. Sub 1, Sub2 represent Submission1 and Submission 2. V1, V2 represent View1 and View2. $g_{11a}$ is observed grade for view1, submission1 with grader a; $g'_{11a}$ is the de-biased grade for view1, submission1 from grader a; $ĝ_{11}$ is consensus grade for view1, submission1; $ĝ_1$ is the overall grade for submission 1. De-biasing process is done based on the view grades collected from each grader. Ga and Gb represent Grader a and Grader b.)

uniform coverage.

3) Collect grades for all submissions: After reviewing each submission, graders need to evaluate and give grades for each view of the submission.

4) Aggregate the grades: The modified Vancouver algorithm which implements a reputation system, which consists of graders' variance and biases, for students is used in this research to aggregate the grades for each view from step 3. This algorithm ensures that higher accuracy leads to higher reputation, and therefore to higher influence on the consensus grades [2].

Each view is considered as one separate aggregation task, and the Vancouver algorithm is applied to iteratively estimate the consensus grade for these views.

5) De-bias each graders' grade for different views: Bias patterns are first identified for each view of every grader using the approach presented in section 3.1. True grades provided by experts are used to compare with the grades given by the graders. The bias patterns are detected from comparisons. Bias is reduced from the view grades afterwards by subtracting min(diff($t_i$, $o_i$)) from the observed grades for a given biased grader. After de-biasing the view grades, the aggregation algorithm in step 4 is applied again on the de-biased data set to get the consensus grades for each view of the submission.

6) Final overall grades for each submission: After applying the aggregation algorithm, each view of every submission will finally get its own consensus grade. The overall grade for each submission is obtained through summation of the view grades.

IV. EXPERIMENTAL ANALYSIS

*4.1 Data Set*

The method is evaluated on synthetic data. In order to compare the proposed approach with the methodology applied in [2] called Vancouver, we used the same synthetic data set. 50 graders and 50 submissions were considered, where each grader was reviewing 6 submissions. To give better understanding of the process, we define 2 different views for the assignment. For each view, we assumed: the true quality $q_i$ of each item i has normal distribution with standard deviation 1. Each grader j had a characteristic variance $v_j$, and we let the grade $q_{ij}$ assigned by j to i be equal to $q_i + \Delta_{ij}$, where $q_i$ is the true quality, and $\Delta_{ij}$ has normal distribution with mean 0 and variance $v_j$. We assumed that the variances $\{v_j\}_{j \in U}$ of the graders were distributed according to a Gamma distribution with scale 0.4, and shape factors k = 2, 3.

*4.2 Models*

Three different models are analyzed, and they are denoted as: AVG, DM1 and DM2.

**AVG** model represents the average model. It just simply averages the grades for each submission received to get the consensus grade. It acts as a baseline model.

**DM1** is the model with different views defined by an expert, but without the step of bias pattern recognition. Thus, there is no de-biasing applied for the observed grades. DM1 is basically Vancouver method in [2], but with extension to different views.

**DM2** model uses both different views of assignments and de-biasing the graders. Different views are defined for the complex assignment before getting the grades from graders. After getting the observed grades, bias patterns are recognized, de-biasing is applied, and finally overall consensus grades are obtained.

*4.3 Evaluation Metrics*

Our goal is to get consensus grades as close as possible to the true grades given by an expert. As a result, coefficient correlation (ρ), standard deviation (σ), and root mean square error (RMSE) are calculated between true grades and consensus grades as the evaluation metrics. The metrics are calculated as follows:

$$\rho_{Y,\hat{Y}} = \frac{COV(Y,\hat{Y})}{\sigma_Y \sigma_{\hat{Y}}} \qquad (6)$$

$$COV(Y,\hat{Y}) = E((Y - \mu_Y)(\hat{Y} - \mu_{\hat{Y}})) \qquad (7)$$

where $\rho_{Y,\hat{Y}}$ is the coefficient correlation of random variables $Y$(true grades) and $\hat{Y}$ (consensus grades), $\sigma_Y, \sigma_{\hat{Y}}$ are the standard deviation of $Y$ and $\hat{Y}$, $\mu_Y, \mu_{\hat{Y}}$ are the mean of $Y$ and $\hat{Y}$, $COV(Y,\hat{Y})$

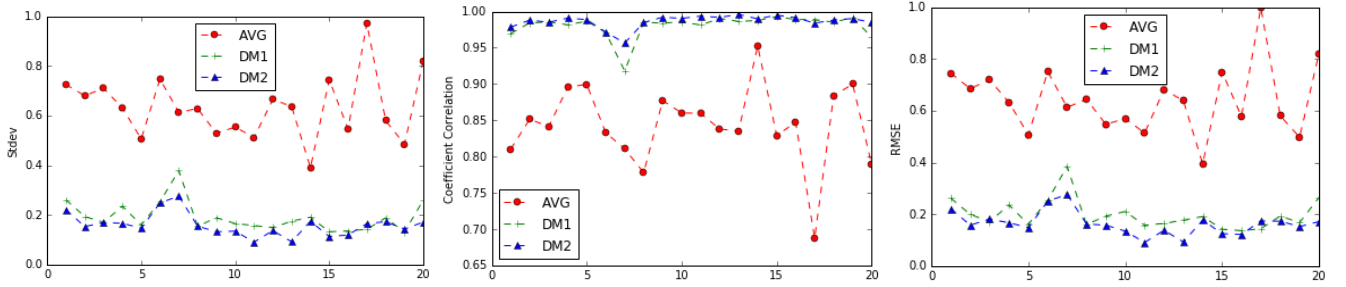
Fig. 3 Standard Deviation, Coefficient Correlation and RMSE for view1 of Synthetic Data for All Three Models for 20 Runs (k=2)

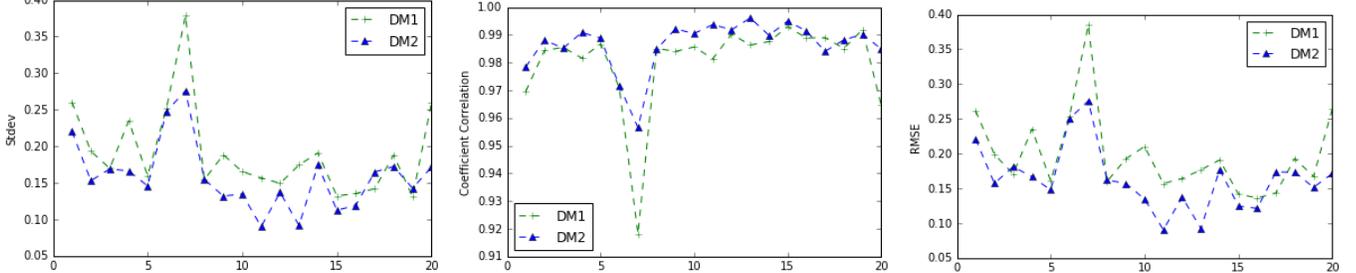
Fig. 4 Standard Deviation, Coefficient Correlation and RMSE for view1 of Synthetic Data for DM1 and DM2 for 20 Runs (k=2)

is the covariance of $Y$ and $\hat{Y}$.

$$RMSE = \sqrt{\frac{\sum_{t=1}^{n}(\hat{y}_t - y)^2}{n}} \quad (8)$$

where $\hat{y}_t$ is the estimated value for true grade and $y$ is the true value.

*4.4 Experimental Results*

The synthetic data is simulated through 100 runs and the coefficient correlation, standard deviation and RMSE are then reported as the average over these 100 runs. Figure 3 shows the evaluation metrics calculated for the view1 of first 20 simulation runs, with shape factor k = 2 of the Gamma distribution, for the three models. To present the difference between DM1 and DM2, we show a more detailed figure in Fig. 4. The result of view2 is quite similar to view1. The results are summarized in Table 1. Specifically, the overall outcome is calculated from aggregation the 100 runs instead of averaging the results from the final views' outcome.

In order to better understand the influence of the bias patterns to the results, we run the following experiment: Increasing the percentage of biased graders (graders who have bias patterns) for each of the view. We increase the number of biased graders for view 1 from 9 to 24, and view 2 from 12 to 28. DM1 and DM2 model are applied to get the consensus grades. We then calculate the metrics for evaluation. Table 2 shows the results.

*4.5 Detailed Experimental Results Analysis and Discussion*

As the results show in Table 1, all three metrics we selected are improved for each view after applying DM1 compared to baseline – AVG model. Thus, the overall results of DM1 model improved 20.3% in contrast to AVG model for correlation when k=2. After de-biasing, which means applying DM2, there is further improvement. As an example, if standards deviation has 70% improvement comparing DM1 with AVG, and 10% improvement comparing DM2 with DM1, then 87% improvement is given comparing DM2 with AVG model. In conclusion, DM2 model gave the best performance for all three evaluation metrics comparing with AVG baseline model and DM1.

Figure 5 and 6 present the percentage of improvement of the results comparing the three different models, which is calculated from Table 1. Figure 5(a) and 6(a) show the enhancement obtained for all three metrics from comparing DM1 model with baseline AVG model, for shape factor equals to 2 and 3 respectively. Figure 5(b) and 6(b) give percentage of improvement of the accuracy after comparing DM2 with DM1 (for k = 2, 3). It could be seen clearly in figure 5(a) and 6(a) that, no matter for each view or overall grades, significant improvement can be obtained by applying DM1 in contrast with AVG model. However, further improvement on the accuracy is able to achieve by using DM2 comparing with DM1. Thus, DM2 gives the best performance.

By defining different views and taking into account graders' bias pattern for each view, we could provide significant gains in accuracy. To prove the importance of de-biasing the graders, we increase the number of biased graders per view in our data set. Table 2 shows the results of three selected metrics results and percentage of improvement comparing DM2 (with de-biasing) with DM1 (without de-biasing). As results presented, the more biased graders, the better improvement we could get by using the DM2 model. The reason is that more graders with a bias pattern means less of them make random error, and as a result, more accurate grades could be obtained after de-biasing the graders.

We compared our method with the Vancouver algorithm in [2] (Alfaro and Shavlovsky, 2014). The Vancouver algorithm is essentially DM1 model, except for that no different views are defined. Thus, the overall results in Table 1 is the approximation of the outcome of Vancouver algorithm. The improvement of metrics of overall grades in Figure 5(b) and 6(b) presents the enhancement on accuracy comparing DM2 model with Vancouver algorithm. Our results show that DM2 yields the better results.

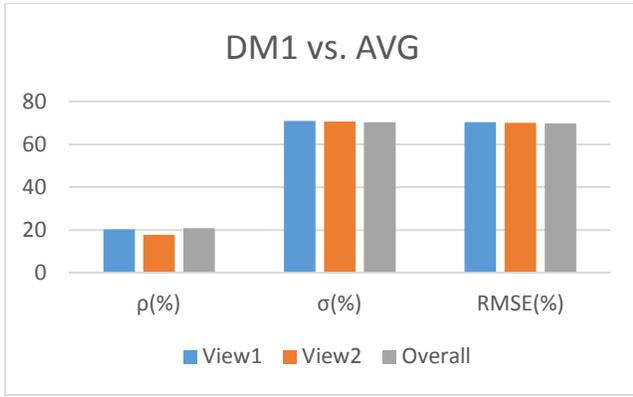

Fig. 5(a) Improvement of Results Comparing DM1 with AVG Model (k=2)

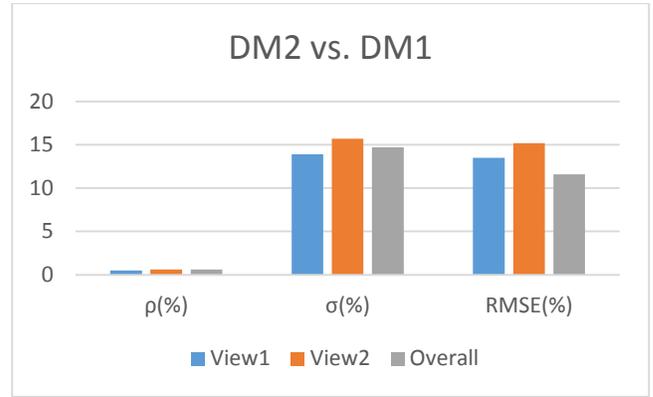

Fig. 5(b) Improvement of Results Comparing DM2 with DM1 Model (k=2)

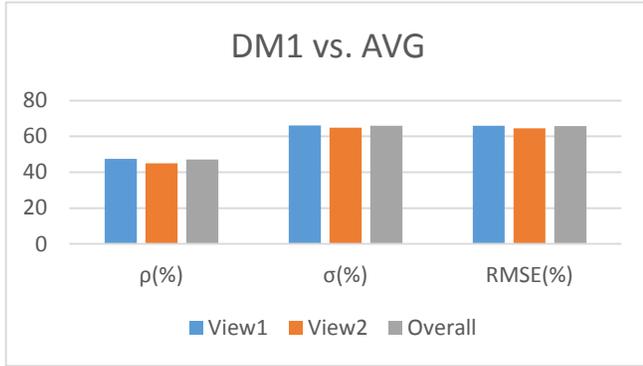

Fig. 6(a) Improvement of Results Comparing DM1 with AVG Model (k=3)

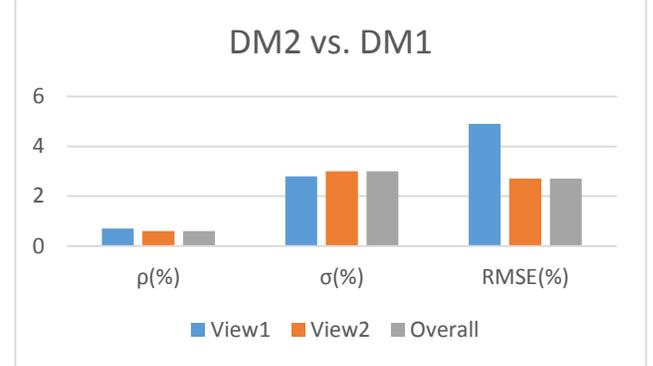

Fig. 6(b) Improvement of Results Comparing DM2 with DM1 Model (k=3)

Table 1 Results for all the models

|         |       | $\rho$ |       | $\sigma$ |       | RMSE  |       |
|---------|-------|-------|-------|-------|-------|-------|-------|
|         |       | k=2   | k=3   | k=2   | k=3   | k=2   | k=3   |
| AVG     | View1 | 0.812 | 0.619 | 0.690 | 1.255 | 0.697 | 1.269 |
|         | View2 | 0.831 | 0.629 | 0.675 | 1.226 | 0.683 | 1.238 |
| DM1     | View1 | 0.977 | 0.913 | 0.201 | 0.425 | 0.207 | 0.431 |
|         | View2 | 0.979 | 0.912 | 0.198 | 0.431 | 0.204 | 0.438 |
| DM2     | View1 | 0.982 | 0.920 | 0.173 | 0.413 | 0.179 | 0.410 |
|         | View2 | 0.985 | 0.918 | 0.167 | 0.418 | 0.173 | 0.426 |
| Overall | AVG   | 0.807 | 0.619 | 0.986 | 1.801 | 0.997 | 1.820 |
|         | DM1   | 0.976 | 0.911 | 0.293 | 0.612 | 0.301 | 0.622 |
|         | DM2   | 0.982 | 0.916 | 0.250 | 0.594 | 0.266 | 0.605 |

Table 2 Results for DM1 and DM2 after increasing the bias percentage

|          |           | View1  |        | View2  |        |
|----------|-----------|--------|--------|--------|--------|
| Num. of biased graders |  | 9 | 24 | 12 | 28 |
| $\rho$   | DM1       | 0.984  | 0.968  | 0.977  | 0.971  |
|          | DM2       | 0.988  | 0.99   | 0.982  | 0.995  |
|          | Impr.(%)  | 0.40%  | 2.27%  | 0.30%  | 2.47%  |
| $\sigma$ | DM1       | 0.177  | 0.223  | 0.225  | 0.246  |
|          | DM2       | 0.149  | 0.137  | 0.193  | 0.0997 |
|          | Impr.(%)  | 15.82% | 38.57% | 14.22% | 59.47% |
| RMSE     | DM1       | 0.179  | 0.438  | 0.225  | 0.451  |
|          | DM2       | 0.153  | 0.154  | 0.197  | 0.1    |
|          | Impr.(%)  | 14.53% | 64.84% | 12.44% | 77.83% |

In research [1], by developing the algorithm for estimating the true grades through statistical models, the authors were able to reduce the RMSE error on their prediction of ground truth by 31% to 33% comparing to baseline model. However, our experimental results show that we are able to reduce RMSE by 67% from 1.820 to 0.605 on overall grades by applying the proposed framework.

## V. CONCLUSIONS AND FUTURE WORKS

In this paper, we presented a novel framework for grading un-decomposable, complex tasks through crowdsourcing. The key innovations include: 1) Extending the Vancouver algorithm to multiple views. 2) De-bias the graders for each view of the task. By defining multiple views for the complex task, uniformity is ensured among the graders. Bias pattern is then detected for each view of the grader, and then we de-bias them in order to improve the grading accuracy. Experimental results indicate that our model DM2 outperform the baseline AVG model and DM1 which doesn't include de-bias process.

To further justify and improve the proposed model, real world complex problems will be formulated, collected, and graded by experienced domain graders. Grading tasks in this work is performed in batch mode. However, a time domain model might reveal the relevance and influences between previous and future grading processes. For example, graders' bias pattern might be reused if some of the graders are persistent for different grading tasks. In addition, the expert background of the graders could also be taken into account while weighting the observed grades for consensus grades. For instance, while incorporating computer science students as graders for grading Java programming assignments, graders who have experience on Java coding are probably more reliable than those who do not. Finally, instead of randomly choosing graders for tasks in this work, we plan to come up with algorithms to take into account the selection of graders with higher accuracy and reliability, as well as optimization of the number of graders for specific complex tasks.